\documentclass{llncs}
\usepackage[numbers,sort&compress,sectionbib]{natbib}
\usepackage{amsmath,graphicx}
\usepackage{amsmath,graphicx}
\usepackage[nolist]{acronym}
\usepackage{tabularx}
\usepackage{color}
\usepackage{multirow}
\usepackage{siunitx}
\usepackage{booktabs} 
 \usepackage[font={small}]{caption, subfig}
\usepackage{tikz}
\newcommand{\imgsize}{0.15}
\newcommand{\tikzbintwo}[1]{
\begin{tikzpicture}
    \node[anchor=south west,inner sep=0] (image) at (0,0) {\includegraphics[width=\imgsize\linewidth]{#1}};
    \begin{scope}[x={(image.south east)},y={(image.north west)}]
       \draw [-stealth, line width=2pt, orange] (.85,.55) -- (.55,.7);  
       \draw [-stealth, line width=2pt, orange] (.2,.92) -- (.4,.87);       
    \end{scope}
\end{tikzpicture}
}

\newcommand{\tikzbinone}[1]{
\begin{tikzpicture}
    \node[anchor=south west,inner sep=0] (image) at (0,0) {\includegraphics[width=\imgsize\linewidth]{#1}};
    \begin{scope}[x={(image.south east)},y={(image.north west)}]
       \draw [-stealth, line width=2pt, orange] (.25,.15) -- (.25,.35);  
    \end{scope}
\end{tikzpicture}
}

\newcommand{\tikzbinthree}[1]{
\begin{tikzpicture}
    \node[anchor=south west,inner sep=0] (image) at (0,0) {\includegraphics[width=\imgsize\linewidth]{#1}};
    \begin{scope}[x={(image.south east)},y={(image.north west)}]
       \draw [-stealth, line width=2pt, orange] (.3,.9) -- (.22,.53);  
       \draw [-stealth, line width=2pt, orange] (.3,.9) -- (.5,.9);  
       \draw [-stealth, line width=2pt, orange] (.3,.9) -- (.7,.5); 
    \end{scope}
\end{tikzpicture}
}

\newcommand{\tikzbinfour}[1]{
\begin{tikzpicture}
    \node[anchor=south west,inner sep=0] (image) at (0,0) {\includegraphics[width=\imgsize\linewidth]{#1}};
    \begin{scope}[x={(image.south east)},y={(image.north west)}]
       \draw [-stealth, line width=2pt, orange] (.6,.7) -- (.42,.5);         
    \end{scope}
\end{tikzpicture}
}

\newcommand{\tikzbinfive}[1]{
\begin{tikzpicture}
    \node[anchor=south west,inner sep=0] (image) at (0,0) {\includegraphics[width=\imgsize\linewidth]{#1}};
    \begin{scope}[x={(image.south east)},y={(image.north west)}]
       \draw [-stealth, line width=2pt, orange] (.54,.65) -- (.34,.6);         
    \end{scope}
\end{tikzpicture}
}

\newcommand{\tikzbinsix}[1]{
\begin{tikzpicture}
    \node[anchor=south west,inner sep=0] (image) at (0,0) {\includegraphics[width=\imgsize\linewidth]{#1}};
    \begin{scope}[x={(image.south east)},y={(image.north west)}]
       \draw [-stealth, line width=2pt, orange] (.5,.55) -- (.45,.71);    
       \draw [-stealth, line width=2pt, orange] (.5,.55) -- (.55,.71);  
       \draw [-stealth, line width=2pt, orange] (.5,.55) -- (.4,.45); 
       \draw [-stealth, line width=2pt, orange] (.5,.55) -- (.6,.45);        
    \end{scope}
\end{tikzpicture}
}

\newcommand{\tikzbinseven}[1]{
\begin{tikzpicture}
    \node[anchor=south west,inner sep=0] (image) at (0,0) {\includegraphics[width=\imgsize\linewidth]{#1}};
    \begin{scope}[x={(image.south east)},y={(image.north west)}]
       \draw [-stealth, line width=2pt, orange] (.5,.78) -- (.5,.58);         
    \end{scope}
\end{tikzpicture}
}

\newcommand{\tikzbineight}[1]{
\begin{tikzpicture}
    \node[anchor=south west,inner sep=0] (image) at (0,0) {\includegraphics[width=\imgsize\linewidth]{#1}};
    \begin{scope}[x={(image.south east)},y={(image.north west)}]
       \draw [-stealth, line width=2pt, orange] (.45,.6) -- (.65,.83);         
    \end{scope}
\end{tikzpicture}
}

\newcommand{\tikzbinnine}[1]{
\begin{tikzpicture}
    \node[anchor=south west,inner sep=0] (image) at (0,0) {\includegraphics[width=\imgsize\linewidth]{#1}};
    \begin{scope}[x={(image.south east)},y={(image.north west)}]       
       \draw [-stealth, line width=2pt, orange] (.45,.6) -- (.7,.73);
    \end{scope}
\end{tikzpicture}
}

\newcommand{\tikzbinten}[1]{
\begin{tikzpicture}
    \node[anchor=south west,inner sep=0] (image) at (0,0) {\includegraphics[width=\imgsize\linewidth]{#1}};
    \begin{scope}[x={(image.south east)},y={(image.north west)}]
       \draw [-stealth, line width=2pt, orange] (.5,.1) -- (.2,.25);
       \draw [-stealth, line width=2pt, orange] (.5,.1) -- (.6,.33);         
    \end{scope}
\end{tikzpicture}
}

\newcolumntype{Y}{>{\centering\arraybackslash}X}
\newcommand{\ie}{\emph{i.e.},}
\newcommand{\eg}{\emph{e.g.},}

\newcommand{\etal}{\emph{et al.}}

\newcommand{\Fig}{Fig.}

\newcommand{\lu}[1]{#1}
\newcommand{\xu}[1]{#1}
\AtBeginDocument{}
\begin{acronym}
\acro{CT}{computed tomography}
\acro{HNN}{holistically-nested network}
\acro{P-HNN}{progressive holistically-nested network}
\acro{ILD}{interstitial lung disease}
\acro{LTRC}{Lung Tissue Research Consortium}
\acro{CNN}{convolutional neural network}
\acro{FCN}{fully convolutional network}
\acro{PLS}{pathological lung segmentation}
\acro{DS}{Dice score}
\acro{CNN}{convolutional neural network}
\acro{ASD}{average surface distance}
\acro{COPD}{chronic obstructive pulmonary disease}
\acro{CV}{cross-validation}
\acro{NIH}{National Institutes of Health}
\acro{UHG}{University Hospitals of Geneva}
\acro{ILD}{interstitial lung disease}
\acro{HU}{Hounsfield unit}
\end{acronym}

\title{Progressive and Multi-Path Holistically Nested Neural Networks for Pathological Lung Segmentation from CT Images}
\author{Adam P. Harrison \and Ziyue Xu\thanks{Corresponding author: ziyue.xu@nih.gov. This work is supported by the Intramural Research Program of the National Institutes of Health, Clinical Center and NIAID and used the computational resources of the NIH HPC Biowulf cluster. (http://hpc.nih.gov). We also thank Nvidia for the donation of a Tesla K40 GPU. This study used data provided by the Lung Tissue Research Consortium supported by the National Heart, Lung, and Blood Institute.}\and Kevin George \and Le Lu \and Ronald M. Summers \and Daniel J. Mollura}

\institute{National Institutes of Health, Bethesda, MD, USA}

\begin{document}
%


\maketitle
\begin{abstract}
\lu{Pathological lung segmentation (PLS) is an important, yet
  challenging, medical image application due to the wide variability of
  pathological lung appearance and shape. Because PLS is often a pre-requisite for
  other imaging analytics, methodological simplicity and generality
  are key factors in usability.} Along those lines, we present a
bottom-up deep-learning based approach that is expressive enough to
handle variations in appearance, while remaining unaffected by any
variations in shape. \lu{We incorporate the deeply supervised learning
  framework, but enhance it with a simple, yet
  effective, progressive multi-path scheme}, which more reliably
merges outputs from different network stages. The result is a deep
model able to produce finer detailed masks, which we call progressive
holistically-nested networks (P-HNNs). \lu{Using extensive cross-validation,
  our method is tested on multi-institutional datasets comprising
  $929$ CT scans} ($848$ publicly available), of \xu{pathological} lungs, reporting mean dice scores of $0.985$ and demonstrating significant qualitative and quantitative improvements over state-of-the art approaches.
\end{abstract}
\begin{keywords}
progressive and multi-path convolutional neural networks, holistically nested neural networks, pathological lung segmentation
\end{keywords}
\section{Introduction}
\label{sec:intro}

Pulmonary diseases are a major source of death and hospitalization worldwide, with \ac{CT} a leading modality for screening~\citep{Mansoor_2015}. Thus, there is great impetus to develop tools for automated detection and diagnosis from \ac{CT}. Reliable \ac{PLS} is a cornerstone of this goal, ensuring that disease detection is not confounded by regions outside the lung~\citep{Mansoor_2015,Baz_2013}. Moreover, \ac{PLS} is also innately useful, \eg{} measuring lung volume. To be effective, \ac{PLS} must handle the wide variability in appearance that abnormalities can cause. 

Leading \ac{PLS} approaches often rely on prior \lu{3D shape or anatomical landmark localization}~\citep{Mansoor_2015,Gill_2014,Sofka_2011}. This \textit{top-down} approach can help delineate regions hard to discriminate based on intensity features alone. However, all shape or localization variations must be accounted for~\citep{Baz_2013,Mansoor_2015}. For instance, the image acting as input in \Fig~\ref{fig:model}, which is of a patient with only one lung, could severely challenge a top-down method that fails to account for this variation. 

In contrast, \xu{with effective image features}, a \textit{bottom-up} approach, \ie{} one that classifies individual pixels or patches, would, in principle, be able to handle \xu{most} \lu{cases with challenging shapes seamlessly}. Existing fully bottom-up \ac{PLS} methods~\citep{Wang_2009,Hosseini_2015} show promise, but the authors of both state their methods can struggle with severe pathologies. Instead, a fruitful direction for \ac{PLS} is deep \acp{FCN}, which currently represents the state-of-the-art within computer vision~\citep{Long_2015,Lin_2016} and medical imaging analysis~\citep{Roth_2016,Ronneberger_2015,Nogues_2016,Cicek_2016,Merkow_2016} for segmentation.  


\lu{Motivated by these developments}, we apply a bottom-up \ac{FCN} approach to \ac{PLS}. Given that \ac{PLS} is often
a first step prior to subsequent analysis, we place a high premium on
simplicity \xu{and robustness}. For this reason, our approach adapts the highly effective, yet straightforward, \ac{HNN} deep-learning
architecture~\citep{Xie_2015}. To overcome issues with \ac{HNN} output ambiguity and the well-known coarsening resolution of \acp{FCN}, we introduce a simple, but surprisingly powerful, multi-path enhancement. Unlike other multi-path
works~\citep{Lin_2016,Cicek_2016,Merkow_2016}, that use complex
coarse-to-fine pathways, we opt for a progressive \lu{constraint on multi-scale pathways} that requires no additional convolutional layers or network parameters. The result is an effective and \lu{uncomplicated} \ac{PLS} solution that we call \acp{P-HNN}. 

%

Focusing on infection-, \ac{ILD}-, and \ac{COPD}-based pathologies, we
test our method on a multi-institute dataset consisting of $929$
challenging thoracic \ac{CT} scans exhibiting a variety of
pathology patterns, including consolidations, infiltrations, fibroses, pleural effusions, lung cavities, and emphysema. This
is the largest analysis to date for assessing \ac{PLS}
performance. Importantly, $846$ of the \ac{CT} scans are publicly
available, allowing future methods to directly compare performance. We
report quantitative five-fold cross-validated metrics, providing a
more realistic picture of our tool's generalizability than prior
work~\citep{Mansoor_2015,Gill_2014,Hosseini_2015,Sofka_2011,Wang_2009}. With
this dataset, we obtain average Dice-scores of $0.985\pm0.011$. We
share our tool online for researchers' use and testing\footnote{https://adampharrison.gitlab.io/p-hnn/}.

\section{Methods}
\label{sec:method}

\begin{figure}
\centering
\includegraphics[scale=.8]{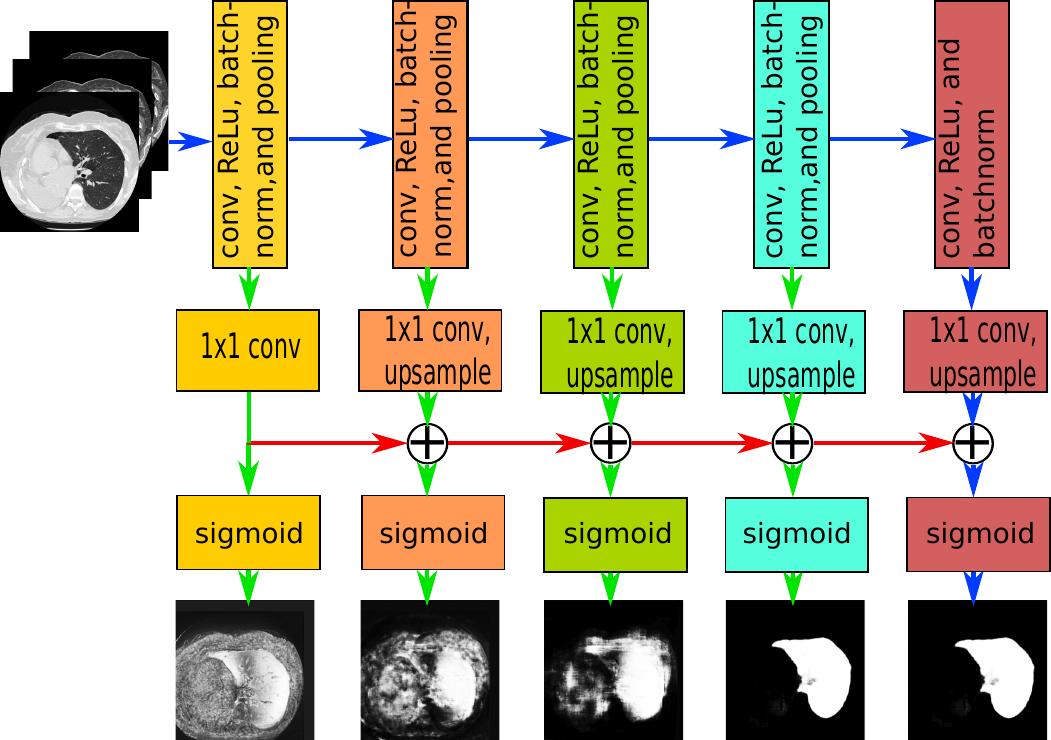}
\caption{\acp{P-HNN} rely on deep supervision and multi-path enhancements, denoted by the green and red arrows, respectively. Blue arrows denote base network components. We use the VGG-16 network~\citep{Simonyan_2015}, minus the fully-connected layers, as our base network. \acp{HNN} only use blue and green pathways, plus a fused output that is not shown. }
\label{fig:model}
\end{figure}

\Fig~\ref{fig:model} illustrates the \ac{P-HNN} model, which progressively refines deeply-supervised mask outputs.
We will first focus on aspects shared with \acp{HNN} and then discuss the progressive multi-path enhancement of \acp{P-HNN}. 

 


To begin, we denote the training data as $S=\{(X_{n},Y_{n}),\,n=1\ldots,N\}$, where $X_{n}$ and $Y_{n}=\{y_{j}^{(n)},\,j=1\ldots,|X_{n}|\},\,y_{j}^{(n)}\in\{0,1\}$ represent the input and binary ground-truth images, respectively. The \ac{HNN} model, originally called holistically-nested edge detection~\citep{Xie_2015}, is a type of \ac{FCN}~\citep{Long_2015}, meaning layer types are limited to convolutional, pooling, and nonlinear activations. \acp{HNN} are built off of a standard \ac{CNN}, \eg{} the VGG-16 model~\citep{Simonyan_2015}, that runs each image through $M$ stages, separated by pooling layers. Unlike the original \ac{HNN}, we use batch normalization after each stage. We denote all network parameters of these standard layers as $\mathbf{W}$.

\acp{HNN} popularized the concept of deep supervision to \acp{FCN}, which is based on the intuition that deeper layers have access to both higher levels of abstraction but coarser levels of scale. As depicted in the green arrows of \Fig~\ref{fig:model}, the novelty of \acp{HNN} is the use of deep supervision to guide the training by computing side-outputs, and their loss, at intermediate stages. \acp{HNN} also merge predictions from different network stages, allowing different levels and scales to contribute to the final result. 

More formally, a set of $1\times 1$ convolutional weights $\mathbf{w}=(\mathbf{w}^{(1)},\ldots\mathbf{w}^{(m)})$ collapses the last activation maps of each stage into an image, \ie{} $a^{(n,m)}_{j}$ for stage $m$, sample $n$, and pixel location $j$. After upsampling to the original resolution, masks at intermediate stages are estimated using
\begin{align}		
	Pr(y_{j}=1|X;\mathbf{W},\mathbf{w}^{(m)})&=\hat{y}^{(n,m)}_{j} \mathrm{,}	\\
	\hat{y}^{(n,m)}_{j} &=\sigma(a^{(n,m)}_{j})\mathrm{,}\\ \label{eqn:hnn_side2}
	\hat{Y}_{n,m}&=\{\hat{y}_{j}^{(n,m)},\,j=1\ldots,|X_{n}|\} \textrm{,}
\end{align}
where $\sigma(.)$ denotes the sigmoid function and $\hat{y}^{(n,m)}_{j}$ and $\hat{Y}_{n,m}$ represent the pixel- and image-level estimates, respectively. We drop $n$ for the remainder of this explanation. 

The cross-entropy loss at each side-output can then be calculated using
\begin{align}
	\ell^{(m)}(\mathbf{W},\mathbf{w}^{(m)})=-\beta\sum_{j\in Y_{+}}\log{\hat{y}^{(m)}_{j}}-(1-\beta)\sum_{j\in Y_{-}}\log\left(1-\hat{y}^{(m)}_{j}\right) \mathrm{,}
\end{align}
where $\beta=\mathit{mean}\left(|Y_{-}|/|Y|\right)$ represents a constant and global class-balancing weight, which we observe provides better \ac{PLS} results than Xie \etal{}'s~\citep{Xie_2015} original image-specific class-balancing scheme. We also prefer to use a sample estimate of the population balance, since we train on an entire training set, and not just on individual images. Not shown in \Fig~\ref{fig:model}, \acp{HNN} also output a final fused probability map based on a learned weighted sum of $\{\hat{Y}_{1},...\hat{Y}_{m}\}$.

%


While highly effective, \acp{HNN} suffer from two issues. The first is inconsistency of the fused output, where certain side outputs on their own can sometimes provide superior results than the final fused output. This is reflected by Xie \etal{}'s use of different outputs depending on the dataset~\citep{Xie_2015}. Ideally, there should be an obvious, and optimal, output to use. 

The second issue is one shared by many early \ac{FCN} solutions---that while deeper stages have greater levels of abstraction, their activations are also coarser in spatial resolution, hampering the capture of fine-level details. This issue is often addressed using multi-path connections~\citep{Lin_2016,Cicek_2016,Merkow_2016} that typically use complex coarse-to-fine pathways to combine activations from earlier stages with later ones, \eg{} the ubiquitous ``U''-like structure~\citep{Lin_2016,Cicek_2016,Merkow_2016}. Multiple convolutional layers~\citep{Cicek_2016,Merkow_2016} or even sub-networks~\citep{Lin_2016} are used to combine activations. Of these, only Merkow \etal{} incorporate both multi-path connections and deep supervision~\citep{Merkow_2016}, but their solution relies on a three-stage training process, in addition to the extra coarse-to-fine pathway.

While these solutions are effective, they require additional parameters totalling nearly the same number as~\citep{Cicek_2016,Merkow_2016}, or more than~\citep{Lin_2016}, the original \ac{FCN} path. Following our philosophy of favoring simplicity, we instead propose a more straightforward \textit{progressive} multi-path connection. As illustrated in \Fig~\ref{fig:model}, we combine activations from the current and previous stages using simple addition prior to applying the sigmoid function. Formally, our model alters the \ac{HNN} formulation by modifying \eqref{eqn:hnn_side2} to
\begin{align}
	\hat{y}^{(m)}_{j} &=\sigma(a^{(m)}_{j}+a^{(m-1)}_{j})\,\,\forall m>1\mathrm{.} \label{eqn:hnn_side2_new}
\end{align}
As activations can exhibit negative or positive values, \eqref{eqn:hnn_side2_new} forces side outputs to improve upon previous outputs, by adding to or subtracting from the corresponding activation. For this reason, we call the model \acfp{P-HNN}. This progressive enhancement allows \acp{P-HNN} to jettison the fused output, avoiding the inconsistent output of \acp{HNN}. Like other multi-path solutions, gradient backpropagation cascades through multiple connections. Unlike other solutions, this enhanced new capability is realized with minimal complexity, requiring fewer parameters than standard \acp{HNN}.

\textbf{Implementation Details:} We train and test our method on 2D axial CT slices, using three windows of $[-1000, 200]$, $[-160, 240]$, and $[-1000,-775]\,$ Hounsfield units to rescale each slice to a $3$-channel $8$-bit image. While 3D \acp{FCN} have been reported~\citep{Cicek_2016,Merkow_2016}, these  rely on \lu{numerous sliding boxes, each with a limited field-of-view. Because lung regions occupy significant portions of a CT scan, large spatial contexts may be needed for accurate segmentation. In addition, due to memory and computational constraints, 3D \acp{CNN} are often less deep and wide than 2D variants.} \lu{Moreover, recent work has demonstrated that 2D \ac{CT} slices are expressive enough for segmenting complex organs~\citep{Ronneberger_2015,Roth_2016,Zhou_2016}. Finally, inter-slice thicknesses of low-dose screening \acp{CT} can range from $5$ to \SI{10}{\milli\metre}. The associated spatial discontinuities could severely challenge 3D-based methods. In contrast, 2D \acp{P-HNN} can work well for any inter-slice thickness.} Thus, we opt for a 2D approach, which remains simple and requires no reduction in \ac{CNN} field-of-view, depth, and width. 

\section{Experiments}
\label{sec:results}

\begin{figure}[t]
\center
\begin{tabular}{@{}c@{}c@{}c@{}c@{}c@{}}
	\tikzbintwo{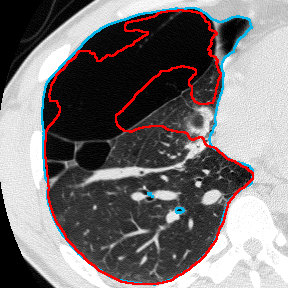} &\tikzbinone{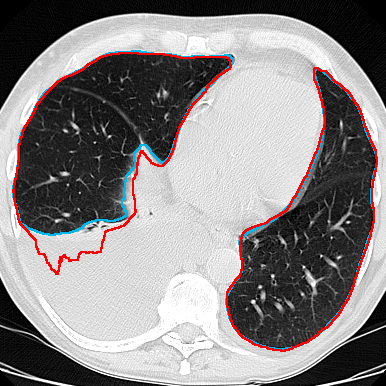} & \tikzbinthree{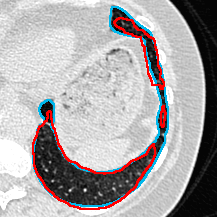} &\tikzbinfour{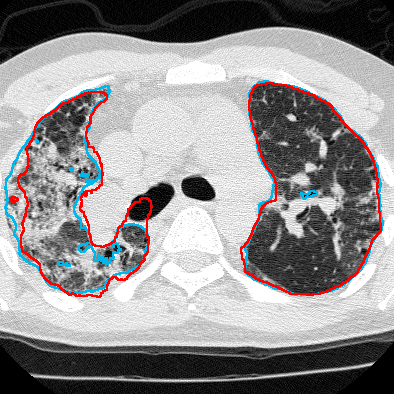} & \tikzbinfive{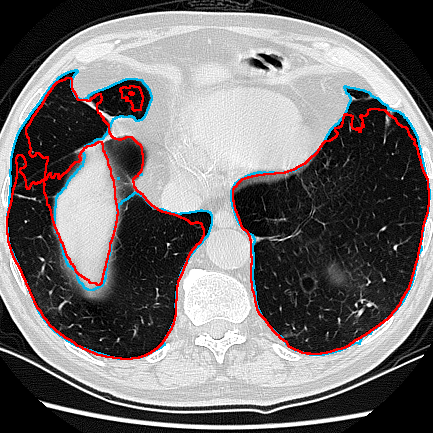} \\
		\tikzbintwo{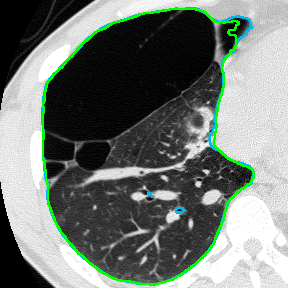} &\tikzbinone{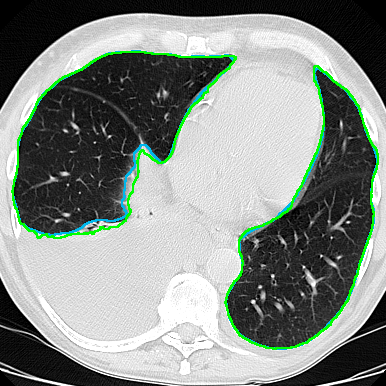} & \tikzbinthree{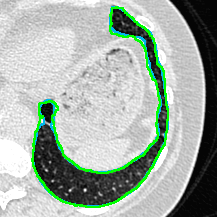} &\tikzbinfour{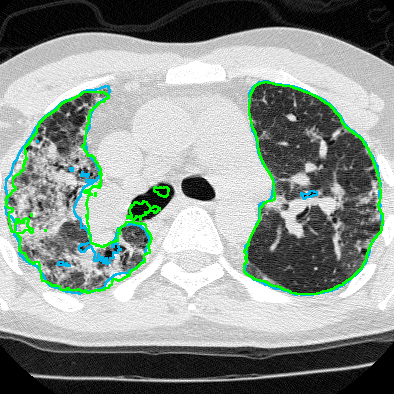} & \tikzbinfive{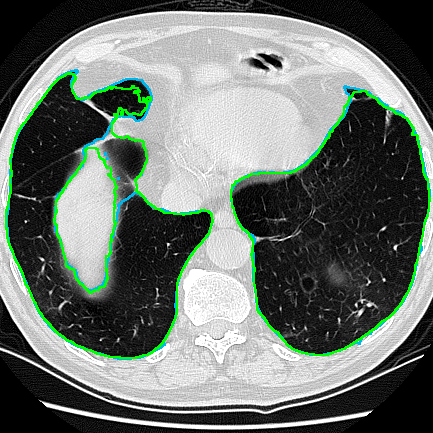} \\
		(a) & (b) & (c) & (d) & (e)
\end{tabular}
\caption{Example masks of \ac{HNN} and \ac{P-HNN}, depicted in red and green, respectively. Ground truth masks are rendered in cyan. (a) \ac{HNN} struggles to segment the pulmonary bullae, whereas \ac{P-HNN} captures it. (b) Part of the pleural effusion is erroneously included by \ac{HNN}, while left out by \ac{P-HNN}. (c) \ac{P-HNN} better captures finer details in the lung mask. (d) In this failure case, both \ac{HNN} and \ac{P-HNN} erroneously include the right main bronchus; however, P-HNN better captures infiltrate regions.  (e) This erroneous ground-truth example, which was filtered out, fails to include a portion of the right lung. Both \ac{HNN} and \ac{P-HNN} capture the region, but \ac{P-HNN} does a much better job of segmenting the rest of the lung.}
\label{fig:example}
\end{figure}

\begin{figure}
\center
\begin{tabular}{@{}c@{}c@{}c@{}c@{}c@{}}
	\tikzbinsix{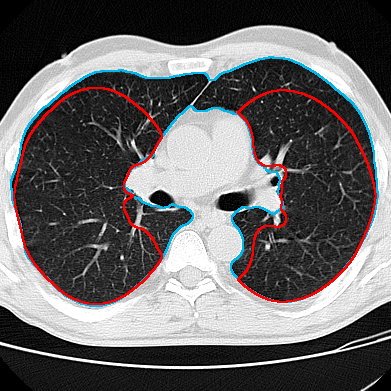} &\tikzbinseven{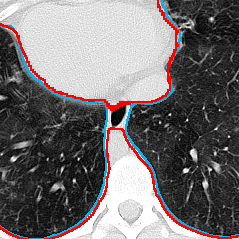} & \tikzbineight{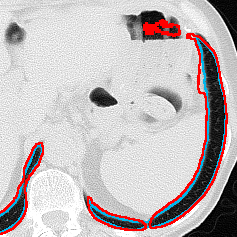} &\tikzbinnine{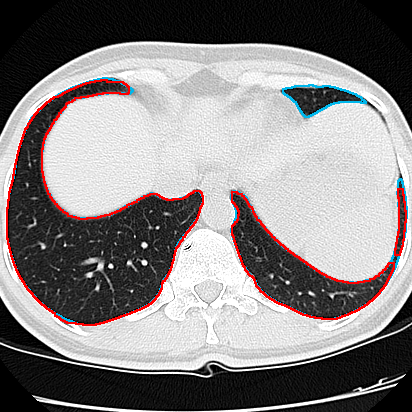} & \tikzbinten{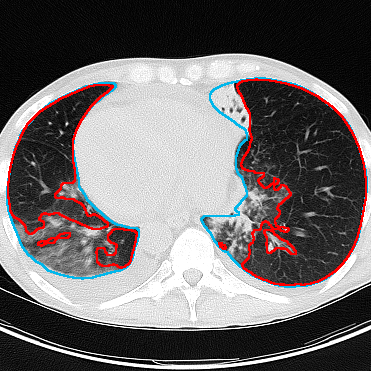} \\
		\tikzbinsix{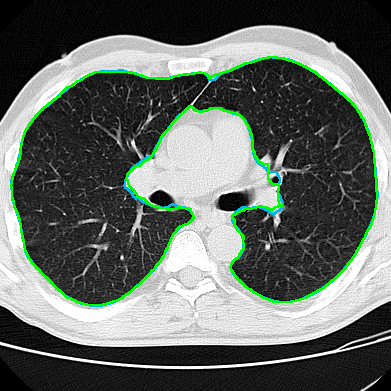} &\tikzbinseven{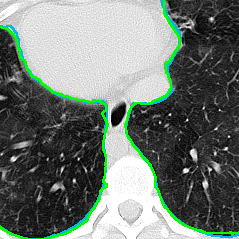} & \tikzbineight{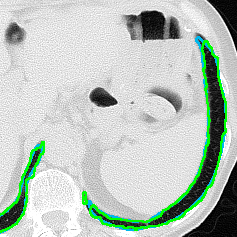}
		 &\tikzbinnine{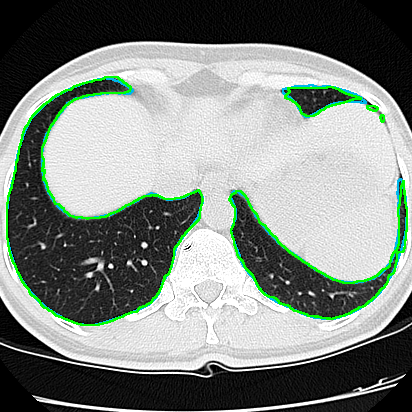} & \tikzbinten{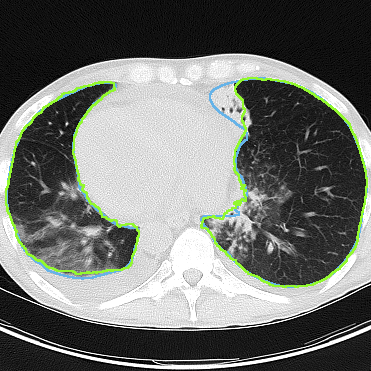} \\
		(a) & (b) & (c) & (d) & (e)
\end{tabular}
\caption{Example masks from Mansoor \etal{}~\citep{Mansoor_2015} and \ac{P-HNN}, depicted in red and green, respectively. Ground truth masks are rendered in cyan. (a) \ac{P-HNN} successfully segments a lung that Mansoor \etal{}'s method is unable to. (b) and (c) Mansoor \etal{}'s method leaks into the esophagus and intestine, respectively. (d) Unlike \ac{P-HNN}, Mansoor \etal{}'s method is unable to fully capture lung field. (e) \ac{P-HNN} better captures regions with ground-glass opacities.}
\label{fig:example2}
\end{figure}

We measure \ac{PLS} performance using multi-institutional data from the
Lung Tissue Research Consortium (LTRC) \ac{ILD}/\ac{COPD} dataset~\citep{Karwoski_2008}, the
\ac{UHG} \ac{ILD}
dataset~\citep{Depeursinge_2012}, and a subset of an infection-based
dataset from the \ac{NIH} Clinical
Center~\citep{Mansoor_2015}\footnote{Due to a data-archiving issue,
  Mansoor \etal{} were only able to share 88 CT scans, and, of those,
  only 47 PLS masks produced by their
  method~\citep{Mansoor_2015}.}. LTRC masks were initially generated
using an automatic method, followed by visual inspection and manual
correction if necessary~\citep{Karwoski_2008}. For all datasets, we
also visually inspect and exclude scan/mask pairs with annotation
errors. This results in $773$, $73$, and $83$ CT scans from the LTRC,
UHG and \ac{NIH} datasets, respectively. 

Using five-fold \ac{CV}, separated at the patient and dataset level, we train on every tenth slice of the LTRC dataset and all slices of the other two. We fine-tuned from an ImageNet pre-trained VGG-16 model~\citep{Simonyan_2015}, halting training after roughly $13.5$ epochs. Validation subsets determined probability-map thresholds. Post-processing simply fills any 3D holes and keeps the largest two connected components if the volume ratio between the two is less than $5$, otherwise only the largest is kept. Depending on the number of slices, our system takes roughly $10-$\SI{30}{\second} to segment one volume using a Tesla K40.

\begin{figure}[t]
\begin{tabular}{cc}
	\includegraphics[width=0.5\linewidth]{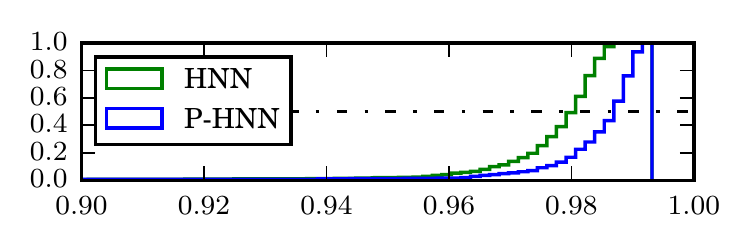}&\includegraphics[width=0.5\linewidth]{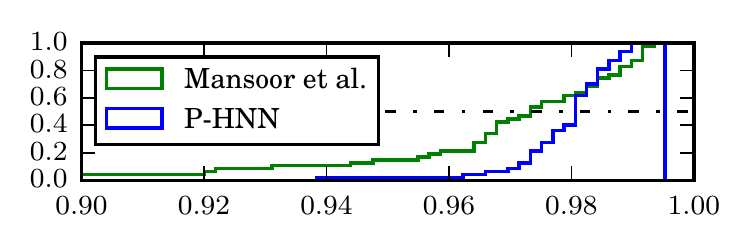}\\
	(a) & (b)
\end{tabular}
\caption{Cumulative histograms of \acp{DS} of \ac{P-HNN} vs. competitors. (a) depict results against standard \ac{HNN} on the entire test set while (b) depicts results against Mansoor \etal{}'s \ac{PLS} tool~\citep{Mansoor_2015} on a subset of $47$ cases with infectious diseases. Differences in score distributions were statistically significant ($p<0.001$) \lu{for both (a) and (b)} using the Wilcox signed-rank test.}
\label{fig:cdf}
\end{figure}

Table~\ref{tbl:ds}(a) depicts
\begin{table}[t]
\caption{Mean \acsp{DS} and \acsp{ASD} and their standard deviation. (a) depicts standard \ac{HNN} and \ac{P-HNN} scores on the entire test dataset. (b) depicts \ac{P-HNN} scores against Mansoor \etal{}'s~\citep{Mansoor_2015} \ac{PLS} method on $47$ volumes from the \ac{NIH} dataset.}
\label{tbl:ds}
\begin{tabular}{cc}
\begin{tiny}
	\begin{tabularx}{.5\linewidth}{YY|Y|Y}
	\textbf{Dataset} & \textbf{Model} & \textbf{DS} & \textbf{ASD (mm)}  \\
	\toprule 
	\multirow{2}{*}{LTRC} & HNN & $0.980\pm0.006$ & $1.045\pm0.365$  \\
	& P-HNN & $0.987\pm0.005$& $0.749\pm0.364$  \\
	\midrule
	\multirow{2}{*}{UHG} & HNN & $0.971\pm0.010$ & $0.527\pm0.287$  \\
	& P-HNN & $0.979\pm0.010$&  $0.361\pm0.319$  \\
	\midrule
	\multirow{2}{*}{\acs{NIH}} & HNN & $0.962\pm0.032$ & $1.695\pm1.239$  \\
	& P-HNN & $0.969\pm0.034$& $1.241\pm1.191$ \\
	\toprule
	\multirow{2}{*}{Total} & HNN & $0.978\pm0.013$ & $1.063\pm0.559$  \\
	& P-HNN & $0.985\pm0.011$& $0.762\pm0.527$ 
	\end{tabularx} \end{tiny}&\begin{tiny}
	\begin{tabularx}{.4\linewidth}{Y|Y|Y}
	\textbf{Model} & \textbf{DS} & \textbf{ASD (mm)}  \\
	\toprule 
	Mansoor \etal{}~\citep{Mansoor_2015} & $0.966\pm0.036$ & $1.216\pm1.491$  \\
	\ac{P-HNN} & $0.980\pm0.009$& $0.827\pm0.436$  \\	
	\end{tabularx}
	 \end{tiny}
	 \\	
	(a) & (b)
\end{tabular}	
\end{table}
the mean 3D \acp{DS} and \acp{ASD} of \ac{HNN} versus \ac{P-HNN}. As can be seen, while \ac{HNN} posts very good \acp{DS} and \acp{ASD} of $0.978$ and $1.063\,$mm, respectively, \ac{P-HNN} is able to outperform it, posting even better values ($p<0.001$ Wilcox signed-rank test) of $0.985$ and $0.762\,$mm, respectively. \Fig~\ref{fig:cdf}(a) depicts cumulative \ac{DS} histograms, visually illustrating the distribution of improvements in segmentation performance. \Fig~\ref{fig:example} depicts selected qualitative examples, demonstrating the effect of these quantitative improvements in \ac{PLS}-mask visual quality and usefulness.

Using $47$ volumes from the \ac{NIH} dataset, we also test against Mansoor \etal{}'s non deep-learning method~\citep{Mansoor_2015}, which produced state-of-the-art performance on challenging and varied infectious disease CT scans. \lu{As Table~\ref{tbl:ds}(b) and \Fig~\ref{fig:cdf}(b) illustrate, \ac{P-HNN} significantly ($p<0.001$) outperforms this previous state-of-the-art approach, producing much higher \acp{DS}. \ac{ASD} scores were also better, but statistical significance was not achieved for this metric. Lastly, as shown in \Fig~\ref{fig:example2}, \ac{P-HNN} generates} PLS masks with considerable qualitative improvements.

\section{Conclusion}
\label{sec:conclusion}

This work introduced \acp{P-HNN}, an \ac{FCN}-based~\cite{Long_2015} deep-learning tool for \ac{PLS} that combines the powerful concepts of deep supervision and multi-path connections. We address the well-known \ac{FCN} coarsening resolution problem using a progressive multi-path enhancement, which, unlike other approaches~\citep{Lin_2016,Cicek_2016,Merkow_2016}, requires no extra parameters over the base \ac{FCN}. When tested on $929$ thoracic CT scans exhibiting infection-, \ac{ILD}-, and \ac{COPD}-based pathologies, the largest evaluation of \ac{PLS} to-date, \ac{P-HNN} \lu{consistently} outperforms \lu{($p<0.001$) standard \ac{HNN}, producing mean \acp{DS} of $0.985\pm0.011$}. \ac{P-HNN} \lu{also provides significantly improved} \ac{PLS} masks  compared against a state-of-the-art tool~\cite{Mansoor_2015}. Thus, \acp{P-HNN} offer a simple, yet highly effective, means to produce robust \ac{PLS} masks. The \ac{P-HNN} model can also be applied to pathological lungs with other morbidities and could provide a straightforward and powerful tool for other segmentation tasks. 


\bibliographystyle{splncs}
\small{
\bibliography{nih_refs1}
}
\end{document}